\documentclass[nohyperref]{article}

\usepackage{microtype}
\usepackage{graphicx}
\usepackage{booktabs} % for professional tables

\usepackage{hyperref}

\usepackage[accepted]{icml2022}

\usepackage{amsmath}
\usepackage{amssymb}
\usepackage{mathtools}
\usepackage{amsthm}
\usepackage{svg}
\usepackage{subcaption}
\usepackage{multirow}
\usepackage{enumitem}

\usepackage[capitalize,noabbrev]{cleveref}

\theoremstyle{plain}

\theoremstyle{definition}

\theoremstyle{remark}

\usepackage[textsize=tiny]{todonotes}

\definecolor{msblue}{HTML}{4572c4}

\icmltitlerunning{OOD-Probe: A Neural Interpretation of Out-of-Domain Generalization}

\begin{document}

\twocolumn[
\icmltitle{OOD-Probe: A Neural Interpretation of Out-of-Domain Generalization}



\icmlsetsymbol{equal}{*}

\begin{icmlauthorlist}
\icmlauthor{Zining Zhu}{uoft,vector}
\icmlauthor{Soroosh Shahtalebi}{vector}
\icmlauthor{Frank Rudzicz}{uoft,vector,uhn}
%\icmlauthor{Firstname4 Lastname4}{sch}
%\icmlauthor{Firstname5 Lastname5}{yyy}
%\icmlauthor{Firstname6 Lastname6}{sch,yyy,comp}
%\icmlauthor{Firstname7 Lastname7}{comp}
%\icmlauthor{Firstname8 Lastname8}{sch}
%\icmlauthor{Firstname8 Lastname8}{yyy,comp}
\end{icmlauthorlist}

\icmlaffiliation{uoft}{University of Toronto}
\icmlaffiliation{vector}{Vector Institute for Artificial Intelligence}
\icmlaffiliation{uhn}{Unity Health Toronto}

\icmlcorrespondingauthor{Zining Zhu}{zining@cs.toronto.edu}
%\icmlcorrespondingauthor{Firstname2 Lastname2}{first2.last2@www.uk}

\icmlkeywords{OOD generalization, probing}

\vskip 0.3in
]



\printAffiliationsAndNotice{\icmlEqualContribution} % otherwise use the standard text.

\begin{abstract}
The ability to generalize out-of-domain (OOD) is an important goal for deep neural network development, and researchers have proposed many high-performing OOD generalization methods from various foundations. While many OOD algorithms perform well in various scenarios, these systems are evaluated as ``black-boxes''. Instead, we propose a flexible framework that evaluates OOD systems with finer granularity using a probing module that predicts the originating domain from intermediate representations. We find that representations always encode some information about the domain. While the layerwise encoding patterns remain largely stable across different OOD algorithms, they vary across the datasets. For example, the information about rotation (on RotatedMNIST) is the most visible on the lower layers, while the information about style (on VLCS and PACS) is the most visible on the middle layers. In addition, the high probing results correlate to the domain generalization performances, leading to further directions in developing OOD generalization systems.
\end{abstract}

\section{Introduction}
Out-of-domain (OOD) generalization is an essential goal in developing deep neural network systems. 
Most existing approaches to developing OOD-generalizable systems follow the invariance principle \citep{arjovsky_invariant_2020}, stating that the data representation should allow the optimal predictor performance to match across environments. Two avenues of work stem from this invariance principle. The first avenue focuses on learning a data representation that remains invariant across environments. This goal can translate to regularization terms that minimize the discrepancy across environments \citep{li_domain_2018}, or auxiliary learning objectives encouraging the representations to be indistinguishable \citep{ganin_DANN_2015,li_CDANN_2018}. %A representation (if treated as a random variable) that is invariant across the environment has little mutual information with the environment. The information-theoretic perspective leads to some learning objectives \citep{ahuja_invariance_2021,ruan_optimal_2021}. 
The second avenue focuses on letting the optimal predictor performance match across environments. To align the predictors, we can align the gradients \citep{shahtalebi_sand-mask_2021,shi_gradient_2021,koyama_when_2021,parascandolo_learning_2020} or prevent the predictors to become overly confident \citep{pezeshki_gradient_2021,wald_calibration_2021}. This avenue requires a diverse collection of environments, which can be implemented by perturbing the domains by group \citep{sagawa_distributionally_2020}, data samples \citep{krueger_out--distribution_2021,yan_mixup_2020}, features \citep{huang_self-challenging_2020}, or some combinations thereof \citep{huang_two_2022}.

Each paper proposes some improvements against multiple previous algorithms and verifies by evaluating accuracy-based scores in novel domains, including the worst group and leave-one-domain-out accuracy \citep{gulrajani_search_2020,ye_ood-bench_2021,hendrycks_many_2021}. While these evaluation protocols provide a holistic perspective for the performance of the networks, they do not reveal the intrinsic mechanisms of generalization. Many questions remain unanswered, including:
\begin{enumerate}
    \item[Q1:] Do the neural networks arrive at invariant representations somewhere in the network? If yes, where?
    \item[Q2:] Do the OOD generalization algorithms encourage the neural networks to learn generalizable representations? If yes, how well do these representations generalize?
\end{enumerate}
We argue that, to answer these questions, we should inspect the intermediate representations of the deep neural networks. To attempt answers, we resort to a tool widely used in the interpretable AI literature -- \textit{probing}. After the deep neural networks are trained, a diagnostic classifier (``probe'') is trained to predict a target from the intermediate representations. A higher probing performance indicates the representation is more relevant to the target \citep{alain_understanding_2017}. Probing analysis reveals many aspects about the intrinsics of deep neural networks (primarily language models), including the encoded semantic knowledge \citep{pavlick_semantic_2022} and linguistic structures \citep{rogers_primer_2020,manning_emergent_2020}. For example, \citet{tenney_bert_2019} found that BERT \citep{devlin-etal-2019-bert}, a Transformer-based language model, automatically forms a pipeline to process textual data in a way that resembles traditional NLP pipelines. The probing results, together with some co-occurrence statistics, can predict the extent to which a feature influences the model's predictions \citep{lovering_predicting_2021}. Probing has demonstrated strong potential for examining the intrinsic mechanisms of deep neural networks.

To answer Q1 and Q2, we set up a framework, OOD-Probe, that attaches an auxiliary probing module to the deep neural network models. OOD-Probe predicts the domain attribute from the intermediate representations.

We apply OOD-Probe to 22 algorithms in DomainBed \citep{gulrajani_search_2020}. Overall, the neural networks do not arrive at truly invariant representations. In addition, the probing results reveal interesting patterns that persist across algorithms and differ across datasets. On RotatedMNIST \citep{ghifary_rotatedMNIST_2015}, the lower layers show easier-to-decode domain attributes. On VLCS \citep{fang_unbiased_2013} and PACS \citep{li_deeper_2017}, the middle layers show the most easily decodable domain attribute. The higher probing performances also correlate well to the OOD generalization performances. In aggregate,  probing performance is predictive of  domain generalization performance and provides evaluations in finer granularity. We call for further attention to the models' intrinsics and discuss how probing can help develop OOD generalization systems. Our analysis codes will be released at \texttt{anonymized-url}.

\section{Method}
\subsection{Problem setting}
We want to set up a neural network system that learns a mapping $f: x\in \mathcal{X}_e \rightarrow y\in \mathcal{Y}_e$, for $e\in \mathcal{E}$. In the ``leave-one-domain-out'' setting of OOD generalization problem, the system $f$ is trained on $M$ training environments $\mathcal{E}_{tr}=\{e_1, e_2, ..., e_{M}\}$ and tested on a novel environment $e_{M+1}$.

\begin{figure}[t]
    \centering
    \includegraphics[width=0.6\linewidth]{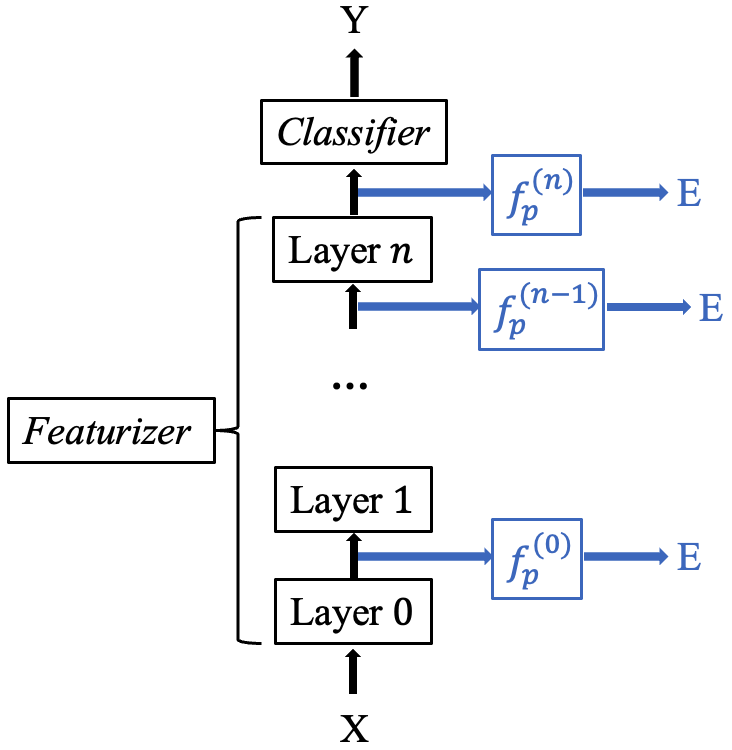}
    \caption{A schematic diagram of the proposed framework. The {\color{msblue}{blue parts}} constitute of the probing module.
    \label{fig:probing_module}}
\end{figure}

The neural network $f$ learns a representation $\Phi$ to represent the random variable $X$ and preserve rich information about $Y$. Usually, the neural networks contain a \textit{Featurizer} (e.g., ResNet \citep{he2016deep} or multilayer CNNs) and a \textit{Classifier} (e.g., Linear), as shown in Figure \ref{fig:probing_module}. Both types of \textit{Featurizer} contain multiple intermediate representations. For multilayer CNNs, we probe the representation for each layer. For ResNet, since the residual connections within each block accelerates the passage of information, we probe the representation from each block.

\subsection{Probing module}
The key component in our neural explanation framework is a probing module, which consists of one or more probes. For a representation $\Phi$ of a trained neural network, we attach a probe, which is a \textit{post-hoc} classifier that predicts a predefined target: $f_p: \Phi \rightarrow T$, where the choice of $T$ depends on the (e.g., linguistic) aspect of the representations to be examined \citep{ettinger_assessing_2018,conneau_senteval_2018}. We set the target $T$ as the environment label $E$. Without loss of generality, one can designate alternative targets as the probing target.

\paragraph{Explaining OOD with probing accuracy}
Many other chocies of $\textrm{Perf}(f_p)$ have been discussed in the probing literature, including the minimum description length of an imaginary channel that transmits information from $\Phi$ to $E$ \citep{voita_information-theoretic_2020}, or some variants of $\Phi - E$ mutual information \citep{pimentel_information-theoretic_2020,hou_birds_2021,hewitt_conditional_2021}. These can be approximated by combinations of cross-entropy losses of probing classifiers. Practically, the precise estimation of information-theoretic terms involving high-dimensional random variables (e.g., $\Phi$) is extremely hard due to limitations of existing estimators \citep{mcallester_formal_2020,song_understanding_2019}. Accuracy is still the most popular choice of $\textrm{Perf}(f_p)$ for classification-type probing problems \citep{belinkov_probing_2021,ravichander_probing_2021,conneau-etal-2018-cram}. In this paper, we also use accuracy. This widely used $\textrm{Perf}(f_p)$ score allows for comparison to the performances of OOD generalization algorithms -- we elaborate the analyses in Section \ref{sec:experiments}.

\section{Experimental setup}
\subsection{Data}
We use four datasets that are widely used in the OOD generalization literature: RotatedMNIST \citep{ghifary_rotatedMNIST_2015}, ColoredMNIST \citep{arjovsky_invariant_2020}, VLCS \citep{fang_unbiased_2013}, and PACS \citep{li_deeper_2017}. These datasets specify classification problems in multiple domains. The variable that denotes the domain, $E$, affects the joint distributions $P(X,Y)$ in distinct manners.

In RotatedMNIST, $E$ specifies the degree of rotation of the handwritten digits. The $E$ in ColoredMNIST corresponds to the proportion of images assigned a colour (red or green). The $f_p: \Phi\rightarrow E$ probing classification performances would be low if the model relies on the digits' shapes but not other factors (e.g., the rotated angle and the colours). 

In PACS, $E$ specifies four distinct styles of images -- photo, art painting, cartoon, and sketch -- all containing images belonging to seven classes: dog, elephant, giraffe, guitar, horse, house, and person. $E$ in VLCS specifies the originating dataset, all of which contain images belonging to five common categories: bird, cat, chair, dog, person. In both PACS and VLCS, the $f_p: \Phi\rightarrow E$ probing classification performance would be low if the representation describes the content but not the styles of the objects.

\subsection{OOD Algorithms}
We attach the probing module to evaluate 22 DomainBed algorithms using their default hyperparameters. The average test-domain out-split accuracies in leave-one-domain-out evaluation are reported in Table \ref{tab:dg_results}.

\subsection{Setting up the probing module}
The most popular type of classifier in the probing literature is a fully-connected linear classifier. Compared to more complicated classifiers, linear classifiers are higher in \textit{selectivity} \citep{hewitt_designing_2019}, i.e., there is a more significant accuracy difference between highly informative representations and less informative ones (specified by control experiments).

Many algorithms in OOD generalization either use multilayer CNNs or ResNet as the \textit{Featurizer}. The immediate input to the \textit{Classifier} is a $D$-dimensional vector. All other hidden representations at layer $i$ follow the shape $(C^{(i)},  H^{(i)}, W^{(i)})$, where $C^{(i)}$, $H^{(i)}$, and $W^{(i)}$ are the number of channels, height, and width at the $i^{\textrm{th}}$ layer, respectively. We flatten the representations to $C^{(i)}\times H^{(i)}\times W^{(i)}$-dimensional vectors and input them to the probes.

Two types of \textit{Featurizer} are used in our experiments: 4-layer CNN for RotatedMNIST and ColoredMNIST, and a ResNet-18 for VLCS and PACS. For the former, we attach a probe at the output of each Conv layer. For the latter, we attach a probe at the output of each block. For both types of networks, we add a classifier to probe the output of the \textit{Featurizer}. 

\subsection{Probing procedure}
We checkpoint the neural networks trained by OOD generalization algorithms and train the probing classifiers from the frozen checkpoints using the same train/validation splits as the OOD training. The probing also follow a ``leave-one-domain-out'' setting: to probe a model $f$ trained on domains $\mathcal{E}_{tr}$ (i.e., leaving out $e_{M+1}$), we train the probing classifier on the same domains to predict $1..M$. 

How many data samples are enough for the probes? We used an off-the-shelf script to recommend probing dataset sizes based on the finite function space bound \citep{zhu_data_2022}. For the probes we set up, we need between 230k to 270k data samples to bound the uncertainty of the probing accuracy within $\pm2\%$, and one epoch (containing $N=5,000$ batches of 64 data samples, totaling 320k data samples) suffices. As an empirical note, most probes' validation accuracies saturate with much fewer than one epoch of samples. Figure \ref{fig:probing_process_coloredmnist} and \ref{fig:probing_process_vlcs} (in the Appendix) provide some examples.

\section{Experiment Results}
\label{sec:experiments}
\subsection{DG algorithms do not remove environment-specific information}
\label{subsec:exp:probing_results}
Figure \ref{fig:probing_accuracy_results} shows the probing accuracy results $\textrm{Perf}(f_p)$, averaged across all domains in the ``leave-one-domain-out'' setting.
To interpret the accuracy results, let us first define a ``dummy classifier'' as one that randomly selects a label with uniform probability. On any dataset with $\textrm{n\_class}$ distinct labels, the dummy classifier is expected to achieve $\frac{1}{\textrm{n\_class}}$ accuracy -- a dummy accuracy. While a worse accuracy is achievable, it is not a meaningful baseline. We set $\frac{1}{\textrm{n\_class}}$ to the lower bounds for the colour bars in Figure \ref{fig:probing_accuracy_results}. 

An accuracy higher than the dummy accuracy by a large margin requires much information from either the data samples or the data distributions. This is observed in all four datasets and most probing tests, indicating that OOD algorithms do not completely remove the environment-specific information. Why is this the case? Here are two alternative hypotheses:
\begin{itemize}[nosep]
    \item While the complete removal of environment-specific information is a desirable goal, the domain-invariant representations are hard to arrive at for current DNN-based learning systems. 
    \item In addition to being invariant among environments, the representations should also be highly beneficial for the original classification problem $f: \Phi \rightarrow Y$. The multiple optimization goals define a complex game where the optimal representations themselves follow a ``trade-off'' among these goals.
\end{itemize}

\subsection{Layerwise patterns across algorithms}
\label{subsec:exp:layerwise_patterns}
Following the rows of Figure \ref{fig:probing_accuracy_results}, consistent patterns are observable. $\textrm{Perf}(f_p)$ decreases as we probe the upper layers for RotatedMNIST. The probing results on ColoredMNIST show a similar pattern, but this trend is not as consistent across algorithms. Note that the probing performances are only slightly above the dummy accuracy at all layers.

On VLCS and PACS, an ``increase-then-decrease'' pattern is visible as we probe into higher layers. The representations from the middle blocks of ResNet-18 encode the domain information in easier-to-decode manners than the remaining blocks. Note that the representation from a block, $\Phi^{(i+1)}$, is \textit{processed} from the representation of the previous block, $\Phi^{(i)}$. Therefore, the data processing inequality guarantees that the information about the environment does not decrease as the block number $i$ increases, i.e., $I(\Phi^{(i)}\,;\,E) \geq I(\Phi^{(i+1)}\,;\,E)$. In other words, the ResNets do not increase the domain information -- they encode the domain information more \textit{linearly} in the middle parts.

Note that the algorithms with similar generalization performance $\textrm{Perf}(f_g)$ could have markedly different probing performances $\textrm{Perf}(f_p)$. For example, CDANN and CORAL achieve $\textrm{Perf}(f_g)$ of 0.97 and 0.98, respectively, on RotatedMNIST. Their probing performances on lower layers are similar, but on the last hidden representation, they get $\textrm{Perf}(f_p)^{(4)}$ of 0.55 and 0.26, respectively.
When the models do not generalize, their representations might encode more or less linearly readable information about the environments. %We discuss further the correlation between the generalization performance and the probing performance later.

\subsection{Layerwise correlations to the OOD performance}
\label{subsec:exp:correlations}
To further understand the utility of the probing performances, we compute the Pearson correlation\footnote{using \texttt{scipy.stats.pearsonr}} of $\textrm{Perf}(f_p)$ and $\textrm{Perf}(f_g)$. Note that the algorithms with values markedly ($\geq 3\sigma$ where $\sigma$ is the std in Table \ref{tab:dg_results}) different from the results report by the DomainBed paper are removed.\footnote{These include: IB\_IRM for RotatedMNIST, CDANN, DANN, IB\_IRM, IRM for PACS.} Table \ref{tab:correlation_results} (in Appendix) shows the results. Additionally, Figure \ref{fig:heatmap_plots} in the Appendix shows the correlation heatmaps. The lower layers of networks on RotatedMNIST show strong correlations, indicating that a linear encoding of the rotation might be beneficial for the generalization. The linear encoding of the styles on PACS shows a similar correlation trend, as reflected by the high correlation values on probes 2 and 3. This trend is slightly different for VLCS, where the Probe\_2 results, which show the highest $\textrm{Perf}(f_p)$, do not correlate as strongly to the OOD performance. 

The above correlations might be affected by some confounding factors, e.g., the inclusion of individual algorithms. To control for this factor, we also compute the correlation w.r.t each algorithm.
Tables \ref{tab:full_correlation_results_rotatedmnist}-\ref{tab:full_correlation_results_PACS} (in Appendix) show the results. In general, $\textrm{Perf}(f_p)$ are positively correlated to $\textrm{Perf}(f_g)$ for RotatedMNIST, ColoredMNIST, and PACS. The correlations are mostly negative for VLCS.

\section{Discussion}
\paragraph{Setting up probing targets}
The environment attribute $E$ is the most convenient target to use, but whether it is the most suitable probing target to set up remains an open question. Our intuition is that a feature $T$ can be a better alternative for the environment attribute $E$ if it can specify the data shift between the environments \textit{concretely}. Otherwise, one can always use the environment attribute. The interpretable NLP literature has many attempts to run carefully-controlled trials that use features to specify the difference between the environments \citep{warstadt_can_2020,mccoy_right_2019,kaushik_learning_2019}. While this approach to constructing datasets is expensive, it allows the designated features to serve as informative probing targets.

\paragraph{Fine-grained evaluations for generalization}
The findings from OOD-Probe open up several possibilities for improving OOD generalization using fine-grained evaluation signals, at least in the following two ways. First, since different modules in the network demonstrate different mechanisms for processing domain-related information, targeted designs may be beneficial. For example, objective functions can be set up to encourage learning linear encoding of target features in lower layers. Second, probing allows a convenient feedback ``dashboard'' for viewing the effects of design choices in building DNN models.

\paragraph{Privacy, fairness, and societal impacts}
The problem of learning and evaluating high-quality representations that are invariant across domains has broad societal impacts. In privacy, the ``domain attribute'' can be the personal identity. This problem can translate to ``removing the speaker attribute from voice while keeping the sounds recognizable'' \citep{tomashenko_voiceprivacy_2022}. If the ``domain attribute'' refers to the demographic property, the problem can be formulated as learning a rich and fair representation \citep{zemel2013learning}. The algorithms that remove protected attributes can have profound long-term impacts on multiple groups \citep{liu_delayed_2018,khani2021removing} -- we defer to \citet{mehrabi_survey_2022} for a summary.

\section{Conclusion}
While OOD generalization is an appealing goal for developing deep neural network systems, their evaluations can be more fine-grained. We propose OOD-Probe, a flexible framework that inspects the neural networks and provides layerwise scores regarding their encoding of the domain attributes. We find patterns that differ across several OOD datasets but remain relatively stable across many algorithms on DomainBed. The probing results show correlation and predictability to the generalization performance, opening up future paths in developing generalizable neural networks.

%\section*{Acknowledgements}


\bibliography{bibliography}
\bibliographystyle{icml2022}

%%%%%%%%%%%%%%%%%%%%%%%%%%%%%%%%%%%%%%%%%%%%%%%%%%%%%%%%%%%%%%%%%%%%%%%%%%%%%%%
%%%%%%%%%%%%%%%%%%%%%%%%%%%%%%%%%%%%%%%%%%%%%%%%%%%%%%%%%%%%%%%%%%%%%%%%%%%%%%%
% APPENDIX
%%%%%%%%%%%%%%%%%%%%%%%%%%%%%%%%%%%%%%%%%%%%%%%%%%%%%%%%%%%%%%%%%%%%%%%%%%%%%%%
%%%%%%%%%%%%%%%%%%%%%%%%%%%%%%%%%%%%%%%%%%%%%%%%%%%%%%%%%%%%%%%%%%%%%%%%%%%%%%%
\newpage
\appendix
\onecolumn
\section*{Appendix}
\section{Experimental Results}

\begin{table}[h]
    \centering
    \caption{Domain generalization leave-one-domain-out accuracy ($\pm$ std)}
    \label{tab:dg_results}
    \resizebox{.9\linewidth}{!}{
    \begin{tabular}{lcccc}
\toprule
Algorithm &    RotatedMNIST &    ColoredMNIST &            VLCS &            PACS \\
\midrule
  ANDMask \citep{parascandolo_learning_2020} & $0.96 \pm 0.03$ & $0.55 \pm 0.30$ & $0.69 \pm 0.18$ & $0.79 \pm 0.10$ \\
      CAD \citep{ruan_optimal_2021} & $0.97 \pm 0.02$ & $0.51 \pm 0.33$ & $0.67 \pm 0.12$ & $0.71 \pm 0.12$ \\
    CDANN \citep{li_CDANN_2018} & $0.97 \pm 0.03$ & $0.54 \pm 0.30$ & $0.53 \pm 0.27$ & $0.18 \pm 0.08$ \\
    CORAL \citep{sun_deep_2016} & $0.98 \pm 0.01$ & $0.51 \pm 0.23$ & $0.73 \pm 0.16$ & $0.83 \pm 0.07$ \\
  CondCAD \citep{ruan_optimal_2021} & $0.97 \pm 0.02$ & $0.49 \pm 0.33$ & $0.66 \pm 0.08$ & $0.77 \pm 0.09$ \\
     DANN \citep{ganin_DANN_2015} & $0.97 \pm 0.03$ & $0.53 \pm 0.33$ & $0.44 \pm 0.26$ & $0.21 \pm 0.05$ \\
      ERM \citep{vapnik1991principles} & $0.98 \pm 0.02$ & $0.52 \pm 0.25$ & $0.74 \pm 0.16$ & $0.83 \pm 0.07$ \\
 GroupDRO \citep{sagawa_distributionally_2020} & $0.98 \pm 0.02$ & $0.55 \pm 0.22$ & $0.73 \pm 0.15$ & $0.83 \pm 0.07$ \\
   IB\_ERM \citep{ahuja_invariance_2021} & $0.98 \pm 0.01$ & $0.50 \pm 0.29$ & $0.74 \pm 0.14$ & $0.79 \pm 0.08$ \\
   IB\_IRM \citep{ahuja_invariance_2021} & $0.20 \pm 0.07$ & $0.51 \pm 0.01$ & $0.49 \pm 0.10$ & $0.13 \pm 0.07$ \\
      IRM \citep{arjovsky_invariant_2020} & $0.70 \pm 0.12$ & $0.56 \pm 0.06$ & $0.49 \pm 0.10$ & $0.16 \pm 0.09$ \\
     MLDG \citep{li_learning_2017} & $0.97 \pm 0.02$ & $0.52 \pm 0.27$ & $0.73 \pm 0.15$ & $0.86 \pm 0.07$ \\
      MMD \citep{li_domain_2018} & $0.98 \pm 0.02$ & $0.37 \pm 0.23$ & $0.72 \pm 0.18$ & $0.82 \pm 0.06$ \\
      MTL \citep{blanchard_MTL_2021} & $0.98 \pm 0.01$ & $0.53 \pm 0.26$ & $0.74 \pm 0.14$ & $0.85 \pm 0.06$ \\
    Mixup \citep{yan_mixup_2020} & $0.97 \pm 0.02$ & $0.52 \pm 0.35$ & $0.76 \pm 0.16$ & $0.83 \pm 0.08$ \\
      RSC \citep{huang_self-challenging_2020} & $0.97 \pm 0.03$ & $0.53 \pm 0.33$ & $0.74 \pm 0.10$ & $0.80 \pm 0.10$ \\
 %SANDMask \citep{shahtalebi_sand-mask_2021} & $0.08 \pm 0.01$ & $0.50 \pm 0.00$ & $0.08 \pm 0.05$ & $0.09 \pm 0.04$ \\
       SD \citep{pezeshki_gradient_2021} & $0.98 \pm 0.02$ & $0.51 \pm 0.25$ & $0.74 \pm 0.17$ & $0.81 \pm 0.09$ \\
   SagNet \citep{nam_reducing_2021}&  $0.98 \pm 0.02$ & $0.52 \pm 0.29$ & $0.75 \pm 0.15$ & $0.83 \pm 0.07$ \\
  SelfReg \citep{kim_selfreg_2021} & $0.98 \pm 0.01$ & $0.51 \pm 0.31$ & $0.76 \pm 0.14$ & $0.82 \pm 0.08$ \\
      TRM \citep{xu_learning_2021} & $0.98 \pm 0.02$ & $0.53 \pm 0.32$ & $0.65 \pm 0.02$ & $0.83 \pm 0.06$ \\
     VREx \citep{krueger_out--distribution_2021} & $0.98 \pm 0.02$ & $0.56 \pm 0.29$ & $0.74 \pm 0.15$ & $0.86 \pm 0.06$ \\
\bottomrule
\end{tabular}
    }
\end{table}

\begin{table}[h]
    \centering
    \caption{Pearson correlations of probing performance $\textrm{Perf}(f_p)$ and domain generalization performance $\textrm{Perf}(f_g)$. $^*$ and $^{**}$ indicate $p<0.05$ and $p<0.01$, respectively.}
    \resizebox{0.8\linewidth}{!}{
    \begin{tabular}{l l l l l l l}
        \toprule 
        Data & Probe 0 & Probe 1 & Probe 2 & Probe 3 & Probe 4 & Probe 5 \\ \midrule 
        RotatedMNIST & \hspace{0.25em}0.7955$^{**}$ & \hspace{0.25em}0.8982$^{**}$ & \hspace{0.25em}0.7451$^{**}$ & \hspace{0.25em}0.4026 & \hspace{0.25em}0.0268 & N/A \\ 
        ColoredMNIST & -0.1859 & -0.0489 & \hspace{0.25em}0.4389$^{*}$ & \hspace{0.25em}0.3646 & \hspace{0.25em}0.2607 & N/A \\
        VLCS & \hspace{0.25em}0.6781$^{**}$ & -0.1489 & \hspace{0.25em}0.1147 & \hspace{0.25em}0.8437$^{**}$ & \hspace{0.25em}0.8434$^{**}$ & \hspace{0.25em}0.8732$^{**}$ \\
        PACS & -0.5936$^{**}$ & \hspace{0.25em}0.6171$^{**}$ & \hspace{0.25em}0.7291$^{**}$ & \hspace{0.25em}0.9703$^{**}$ & \hspace{0.25em}0.8202$^{**}$ & \hspace{0.25em}0.6592$^{**}$ \\ \bottomrule 
    \end{tabular}}
    \label{tab:correlation_results}
\end{table}

\begin{figure}[h]
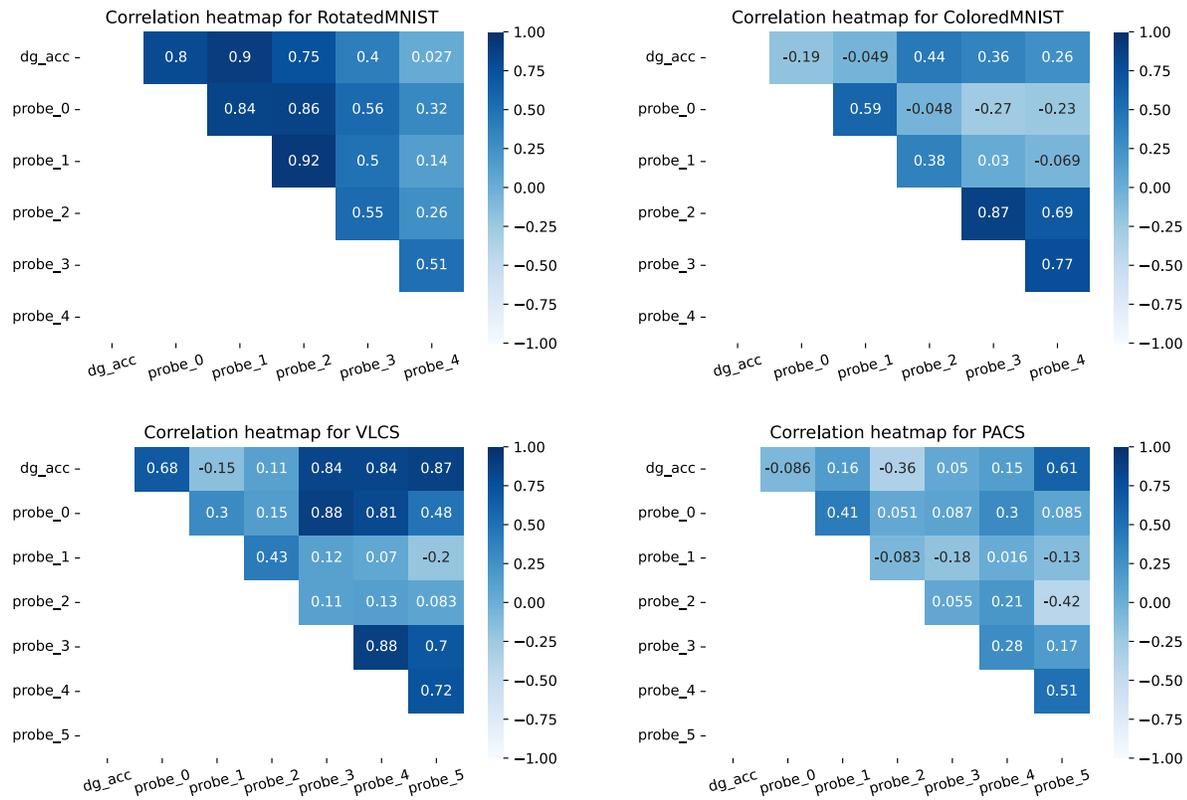

    \centering
    \includesvg[width=.48\linewidth]{figs/heatmap_rotatedmnist.svg}
    \includesvg[width=.48\linewidth]{figs/heatmap_coloredmnist.svg}
    \includesvg[width=.48\linewidth]{figs/heatmap_VLCS.svg}
    \includesvg[width=.48\linewidth]{figs/heatmap_PACS.svg}
    \caption{Correlation within the probing results, and those between the probing performances and the OOD generalization performances.}
    \label{fig:heatmap_plots}
\end{figure}

\begin{figure}
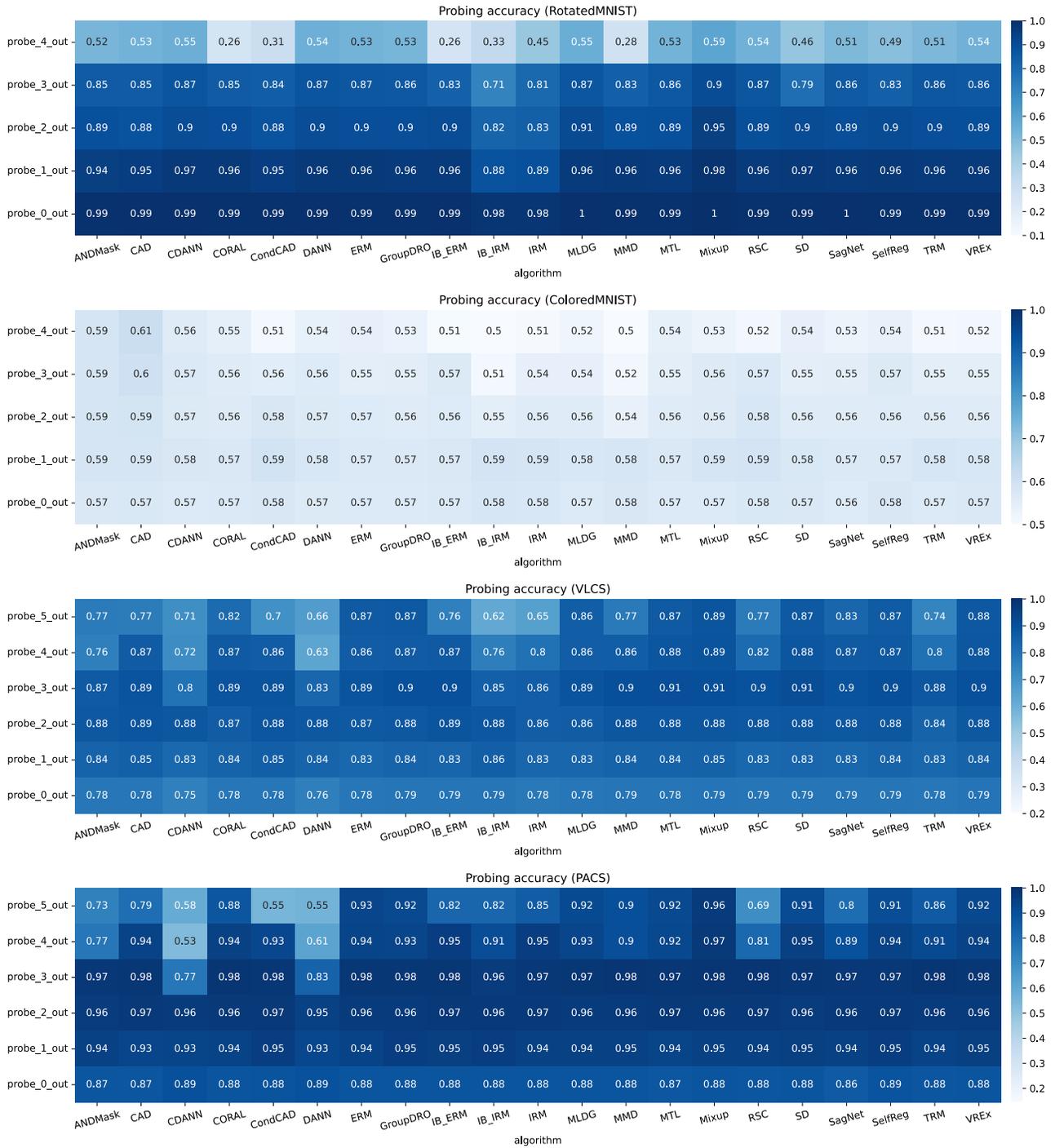

    \centering
    \includesvg[width=\linewidth]{figs/accuracy_RotatedMNIST.svg}
    \includesvg[width=\linewidth]{figs/accuracy_ColoredMNIST.svg}
    \includesvg[width=\linewidth]{figs/accuracy_VLCS.svg}
    \includesvg[width=\linewidth]{figs/accuracy_PACS.svg}
    \caption{Probing accuracies on four datasets. The color bars scale from $\frac{1}{\textrm{n\_class}}$ to $1.00$.}
    \label{fig:probing_accuracy_results}
\end{figure}

\begin{figure}
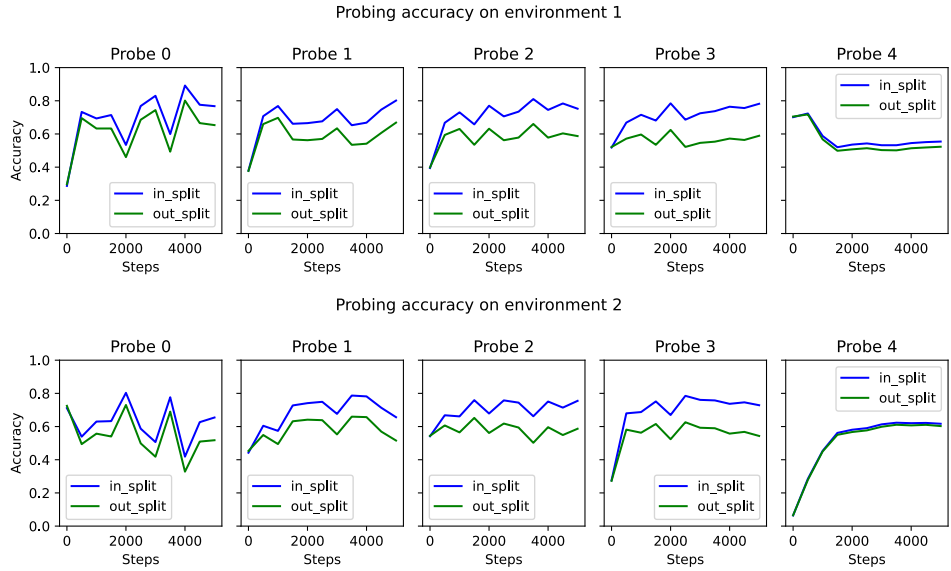

    \centering
    \includesvg[width=.75\linewidth]{figs/probing_process/ERM_ColoredMNIST_testenv0_1.svg}
    \includesvg[width=.75\linewidth]{figs/probing_process/ERM_ColoredMNIST_testenv0_2.svg}
    \caption{Probing accuracy on a checkpoint of ERM on ColoredMNIST. The test environment of ERM is 0, which is left out by the probe.}
    \label{fig:probing_process_coloredmnist}
\end{figure}

\begin{figure}
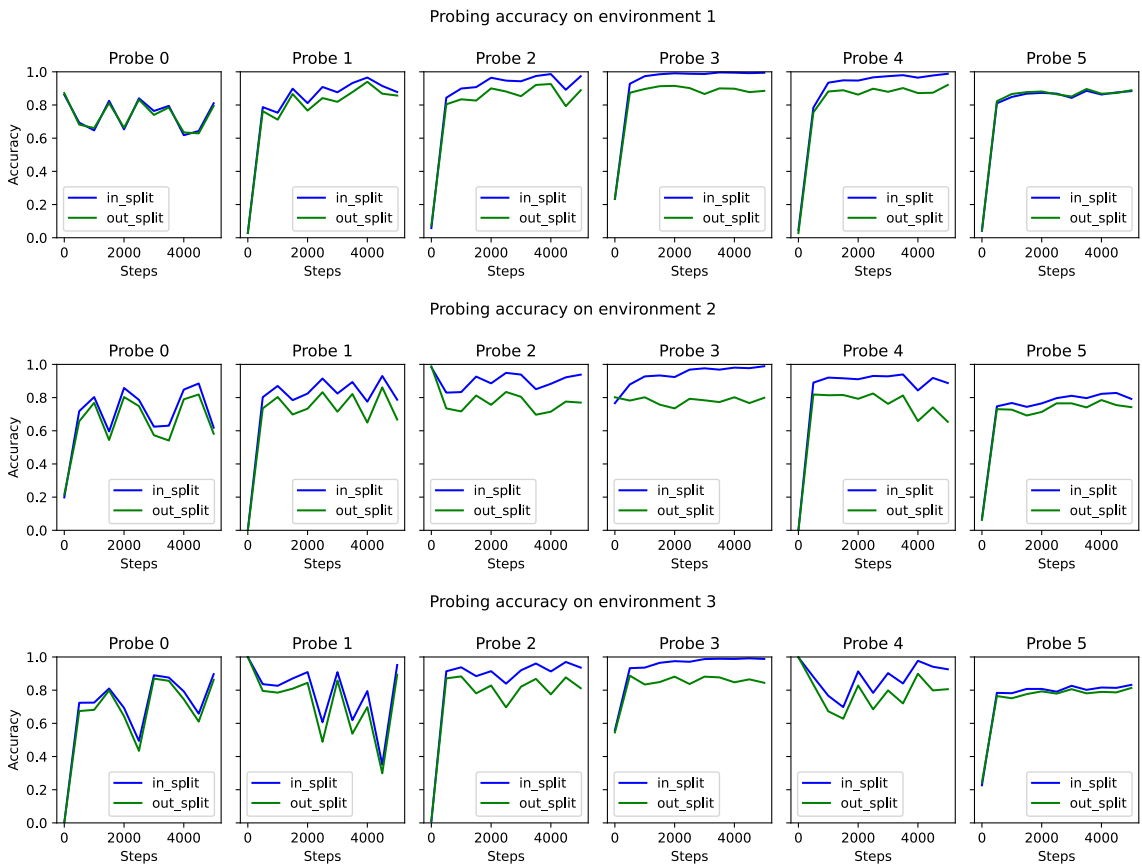

    \centering
    \includesvg[width=.9\linewidth]{figs/probing_process/ERM_VLCS_testenv0_1.svg}
    \includesvg[width=.9\linewidth]{figs/probing_process/ERM_VLCS_testenv0_2.svg}
    \includesvg[width=.9\linewidth]{figs/probing_process/ERM_VLCS_testenv0_3.svg}
    \caption{Probing accuracy on a checkpoint of ERM on VLCS with test environment set to 0.}
    \label{fig:probing_process_vlcs}
\end{figure}

\begin{figure}
    \centering
    \begin{subfigure}[b]{.9\linewidth}
        \includesvg[width=\linewidth]{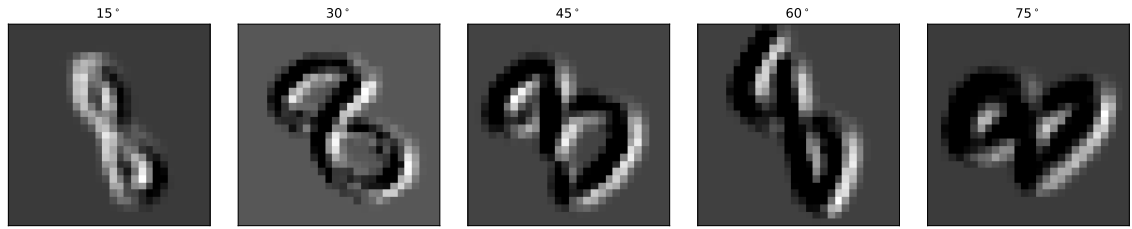}
        \caption{Probe 0 ($28\times 28$). It is relatively easy to tell apart both the originating domain (rotated angles) and the contents.}
    \end{subfigure}
    \begin{subfigure}[b]{.9\linewidth}
        \includesvg[width=\linewidth]{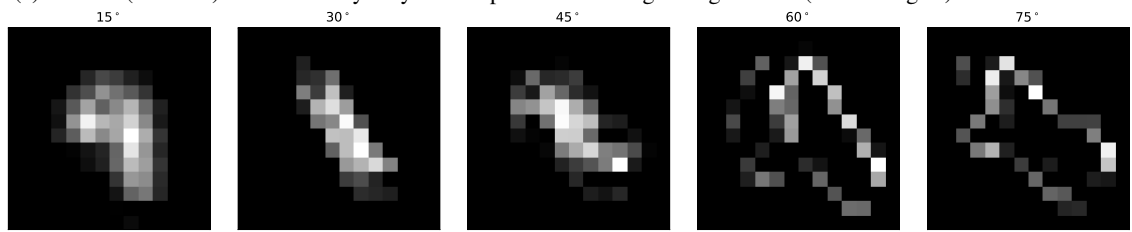}
        \caption{Probe 1 ($14\times 14$). The contents are less obvious, but the degrees of rotations are still visible.}
    \end{subfigure}
    \begin{subfigure}[b]{.9\linewidth}
        \includesvg[width=\linewidth]{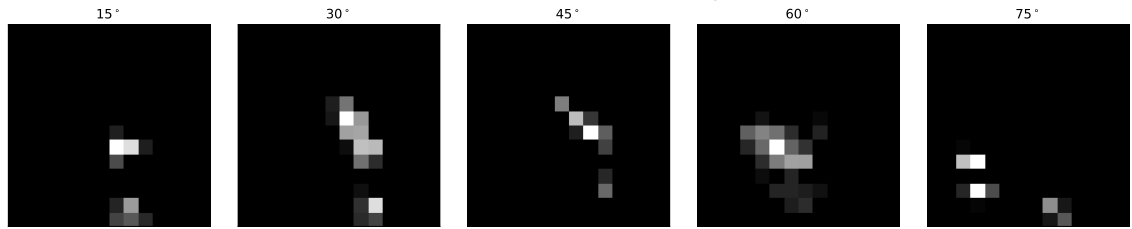}
        \caption{Probe 2 ($14\times 14$). The representations appear more abstract than the previous layers, but we can still guess the rotation angles to some extent.}
    \end{subfigure}
    \begin{subfigure}[b]{.9\linewidth}
        \includesvg[width=\linewidth]{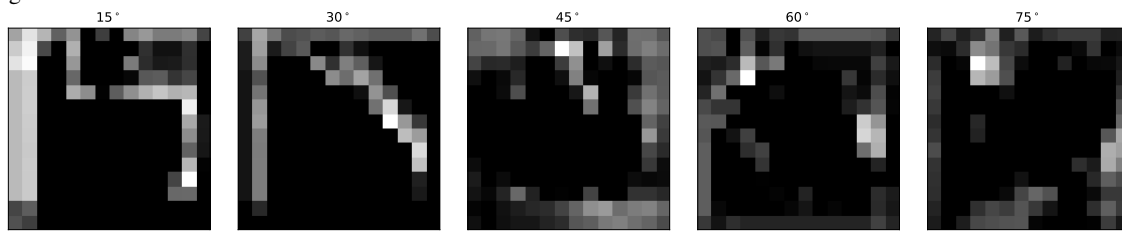}
        \caption{Probe 3 ($14\times 14$). Neither the digit contents nor the rotation angles remain intelligible. However, the probing classifiers can reach .75 accuracy.}
    \end{subfigure}
    
    \caption{Intermediate representations of Rotated MNIST (trained by ERM) as the inputs of the first four probes. The fifth probe takes one-dimensional inputs, whereas only two-dimensional representations are intelligible via plotting. Each representation contains 64, 128, 128, and 128 channels, respectively. Only one channel is (randomly) sampled to visualize.}
    \label{fig:visualization_representation_rotated_mnist}
\end{figure}

\begin{table}[h]
    \centering
    \caption{Correlations between probing results and generalization accuracies, on RotatedMNIST}
    \resizebox{0.8\linewidth}{!}{
    \begin{tabular}{lllllll}
\toprule
Algorithm & Probe\_0 & Probe\_1 & Probe\_2 & Probe\_3 & Probe\_4 & Probe\_5 \\
\midrule
ANDMask   &       \hspace{0.25em}0.2881 &  \hspace{0.25em}0.9437$^{***}$ &  \hspace{0.25em}0.9572$^{***}$ &  \hspace{0.25em}0.9483$^{***}$ &  \hspace{0.25em}0.9964$^{***}$ &  N/A \\
CAD       &          \hspace{0.25em}0.6126 &  \hspace{0.25em}0.9242$^{***}$ &    \hspace{0.25em}0.856$^{**}$ &   \hspace{0.25em}0.8381$^{**}$ &   \hspace{0.25em}0.8411$^{**}$ &  N/A \\
CDANN     &        \hspace{0.25em}0.8551$^{**}$ &     \hspace{0.25em}0.83$^{**}$ &  \hspace{0.25em}0.9461$^{***}$ &  \hspace{0.25em}0.9482$^{***}$ &  \hspace{0.25em}0.9499$^{***}$ &  N/A \\
CORAL     &        \hspace{0.25em}0.8307$^{**}$ &   \hspace{0.25em}0.8858$^{**}$ &  \hspace{0.25em}0.9213$^{***}$ &  \hspace{0.25em}0.9198$^{***}$ &     \hspace{0.25em}0.0953 &  N/A \\
CondCAD   &       \hspace{0.25em}0.9311$^{***}$ &  \hspace{0.25em}0.9198$^{***}$ &  \hspace{0.25em}0.9481$^{***}$ &   \hspace{0.25em}0.8669$^{**}$ &     \hspace{0.25em}0.5268 &  N/A \\
DANN      &          \hspace{0.25em}0.6012 &   \hspace{0.25em}0.8587$^{**}$ &  \hspace{0.25em}0.9222$^{***}$ &   \hspace{0.25em}0.921$^{***}$ &  \hspace{0.25em}0.9667$^{***}$ &  N/A \\
ERM       &        \hspace{0.25em}0.8494$^{**}$ &   \hspace{0.25em}0.8367$^{**}$ &    \hspace{0.25em}0.8013$^{*}$ &   \hspace{0.25em}0.8496$^{**}$ &  \hspace{0.25em}0.9335$^{***}$ &  N/A \\
GroupDRO  &          \hspace{0.25em}0.7032 &   \hspace{0.25em}0.8758$^{**}$ &   \hspace{0.25em}0.8952$^{**}$ &   \hspace{0.25em}0.8978$^{**}$ &   \hspace{0.25em}0.9108$^{**}$ &  N/A \\
IB\_ERM    &          \hspace{0.25em}0.6786 &     \hspace{0.25em}0.6929 &   \hspace{0.25em}0.8788$^{**}$ &    \hspace{0.25em}0.7546$^{*}$ &     \hspace{0.25em}0.3893 &  N/A \\
IB\_IRM    &          \hspace{0.25em}0.4827 &     \hspace{0.25em}0.6954 &      \hspace{0.25em}0.724 &     \hspace{0.25em}0.4284 &     \hspace{0.25em}0.0707 &  N/A \\
IRM       &          \hspace{0.25em}0.4979 &   \hspace{0.25em}0.8738$^{**}$ &   \hspace{0.25em}0.8717$^{**}$ &  \hspace{0.25em}0.9316$^{***}$ &   \hspace{0.25em}0.8896$^{**}$ &  N/A \\
MLDG      &          \hspace{0.25em}0.7268 &  \hspace{0.25em}0.9349$^{***}$ &  \hspace{0.25em}0.9366$^{***}$ &  \hspace{0.25em}0.9632$^{***}$ &   \hspace{0.25em}0.8557$^{**}$ &  N/A \\
MMD       &         \hspace{0.25em}0.7999$^{*}$ &   \hspace{0.25em}0.8775$^{**}$ &   \hspace{0.25em}0.933$^{***}$ &  \hspace{0.25em}0.9493$^{***}$ &     \hspace{0.25em}0.6374 &  N/A \\
MTL       &        \hspace{0.25em}0.8572$^{**}$ &    \hspace{0.25em}0.8059$^{*}$ &   \hspace{0.25em}0.8892$^{**}$ &   \hspace{0.25em}0.8879$^{**}$ &  \hspace{0.25em}0.9716$^{***}$ &  N/A \\
Mixup     &          \hspace{0.25em}0.5907 &   \hspace{0.25em}0.8608$^{**}$ &     \hspace{0.25em}0.6677 &    \hspace{0.25em}0.7965$^{*}$ &  \hspace{0.25em}0.9684$^{***}$ &  N/A \\
RSC       &          \hspace{0.25em}0.7001 &  \hspace{0.25em}0.9388$^{***}$ &   \hspace{0.25em}0.9051$^{**}$ &   \hspace{0.25em}0.9155$^{**}$ &  \hspace{0.25em}0.9747$^{***}$ &  N/A \\
%SANDMask  &                       -0.7944$^{*}$ &                   -0.7202 &                   -0.6037 &                   -0.4775 &                   -0.4594 &  N/A \\
SD        &         \hspace{0.25em}0.917$^{**}$ &    \hspace{0.25em}0.913$^{**}$ &  \hspace{0.25em}0.9409$^{***}$ &  \hspace{0.25em}0.9573$^{***}$ &    \hspace{0.25em}0.7983$^{*}$ &  N/A \\
SagNet    &         \hspace{0.25em}0.7862$^{*}$ &   \hspace{0.25em}0.8598$^{**}$ &   \hspace{0.25em}0.9141$^{**}$ &   \hspace{0.25em}0.9094$^{**}$ &  \hspace{0.25em}0.9466$^{***}$ &  N/A \\
SelfReg   &        \hspace{0.25em}0.9142$^{**}$ &   \hspace{0.25em}0.8514$^{**}$ &  \hspace{0.25em}0.9222$^{***}$ &   \hspace{0.25em}0.9134$^{**}$ &    \hspace{0.25em}0.7575$^{*}$ &  N/A \\
TRM       &          \hspace{0.25em}0.6009 &   \hspace{0.25em}0.8282$^{**}$ &   \hspace{0.25em}0.8852$^{**}$ &  \hspace{0.25em}0.9188$^{***}$ &   \hspace{0.25em}0.9054$^{**}$ &  N/A \\
VREx      &          \hspace{0.25em}0.6077 &   \hspace{0.25em}0.8968$^{**}$ &   \hspace{0.25em}0.8604$^{**}$ &   \hspace{0.25em}0.8569$^{**}$ &  \hspace{0.25em}0.9847$^{***}$ &  N/A \\
\bottomrule
\end{tabular}
    }
    \label{tab:full_correlation_results_rotatedmnist}
\end{table}

\begin{table}[h]
    \centering
    \caption{Correlations between probing results and generalization accuracies, on ColoredMNIST}
    \resizebox{0.8\linewidth}{!}{
    \begin{tabular}{lllllll}
\toprule
Algorithm & Probe\_0 & Probe\_1 & Probe\_2 & Probe\_3 & Probe\_4 & Probe\_5 \\
\midrule
ANDMask   &     \hspace{0.25em}0.9697 &     \hspace{0.25em}0.986 &   \hspace{0.25em}0.9852 &   \hspace{0.25em}0.9888$^{*}$ &  \hspace{0.25em}0.9981$^{**}$ &  N/A \\
CAD       &        \hspace{0.25em}0.9747 &    \hspace{0.25em}0.9725 &   \hspace{0.25em}0.9585 &    \hspace{0.25em}0.9872 &   \hspace{0.25em}0.9896$^{*}$ &  N/A \\
CDANN     &        \hspace{0.25em}0.9585 &    \hspace{0.25em}0.9726 &   \hspace{0.25em}0.9725 &  \hspace{0.25em}0.9981$^{**}$ &    \hspace{0.25em}0.9623 &  N/A \\
CORAL     &       \hspace{0.25em}0.9903$^{*}$ &  \hspace{0.25em}0.9992$^{**}$ &  \hspace{0.25em}0.9968$^{*}$ &  \hspace{0.25em}0.9993$^{**}$ &   \hspace{0.25em}0.9908$^{*}$ &  N/A \\
CondCAD   &        \hspace{0.25em}0.9827 &  \hspace{0.25em}0.9997$^{**}$ &  \hspace{0.25em}0.9928$^{*}$ &    \hspace{0.25em}0.6784 &    \hspace{0.25em}0.2138 &  N/A \\
DANN      &        \hspace{0.25em}0.9263 &    \hspace{0.25em}0.9764 &   \hspace{0.25em}0.9781 &    \hspace{0.25em}0.9818 &    \hspace{0.25em}0.7047 &  N/A \\
ERM       &        \hspace{0.25em}0.9762 &    \hspace{0.25em}0.9873 &  \hspace{0.25em}0.9933$^{*}$ &    \hspace{0.25em}0.9874 &   \hspace{0.25em}0.9907$^{*}$ &  N/A \\
GroupDRO  &        \hspace{0.25em}0.9788 &   \hspace{0.25em}0.9885$^{*}$ &   \hspace{0.25em}0.9858 &    \hspace{0.25em}0.9776 &    \hspace{0.25em}0.9818 &  N/A \\
IB\_ERM    &       \hspace{0.25em}0.9879$^{*}$ &    \hspace{0.25em}0.9761 &  \hspace{0.25em}0.9955$^{*}$ &    \hspace{0.25em}0.9847 &    \hspace{0.25em}0.9747 &  N/A \\
IB\_IRM    &        \hspace{0.25em}0.2722 &    \hspace{0.25em}0.2686 &   \hspace{0.25em}0.1684 &    \hspace{0.25em}0.1637 &    \hspace{0.25em}0.7736 &  N/A \\
IRM       &         \hspace{0.25em}0.945 &    \hspace{0.25em}0.9233 &    \hspace{0.25em}0.983 &  \hspace{0.25em}0.9998$^{**}$ &    \hspace{0.25em}0.9792 &  N/A \\
MLDG      &        \hspace{0.25em}0.9782 &  \hspace{0.25em}0.9993$^{**}$ &  \hspace{0.25em}0.9953$^{*}$ &   \hspace{0.25em}0.9908$^{*}$ &    \hspace{0.25em}0.9861 &  N/A \\
MMD       &        \hspace{0.25em}0.9185 &    \hspace{0.25em}0.995$^{*}$ &   \hspace{0.25em}0.9824 &    \hspace{0.25em}0.7451 &    \hspace{0.25em}1.0$^{***}$ &  N/A \\
MTL       &        \hspace{0.25em}0.9722 &   \hspace{0.25em}0.9907$^{*}$ &   \hspace{0.25em}0.9868 &   \hspace{0.25em}0.9881$^{*}$ &   \hspace{0.25em}0.999$^{**}$ &  N/A \\
Mixup     &         \hspace{0.25em}0.971 &    \hspace{0.25em}0.9871 &   \hspace{0.25em}0.9861 &  \hspace{0.25em}0.9983$^{**}$ &    \hspace{0.25em}0.7062 &  N/A \\
RSC       &        \hspace{0.25em}0.9578 &    \hspace{0.25em}0.9568 &    \hspace{0.25em}0.979 &   \hspace{0.25em}0.9941$^{*}$ &                  -0.0878 &  N/A \\
%SANDMask  &                       -0.649 &                  -0.6457 &                 -0.6068 &                  -0.6214 &                  -0.7239 &  N/A \\
SD        &        \hspace{0.25em}0.9787 &   \hspace{0.25em}0.9948$^{*}$ &   \hspace{0.25em}0.9837 &    \hspace{0.25em}1.0$^{***}$ &    \hspace{0.25em}0.9861 &  N/A \\
SagNet    &         \hspace{0.25em}0.982 &   \hspace{0.25em}0.9935$^{*}$ &   \hspace{0.25em}0.9822 &    \hspace{0.25em}0.9837 &    \hspace{0.25em}0.9255 &  N/A \\
SelfReg   &        \hspace{0.25em}0.9868 &   \hspace{0.25em}0.9951$^{*}$ &  \hspace{0.25em}0.9892$^{*}$ &    \hspace{0.25em}0.9777 &    \hspace{0.25em}0.7666 &  N/A \\
TRM       &        \hspace{0.25em}0.991$^{*}$ &  \hspace{0.25em}0.9978$^{**}$ &  \hspace{0.25em}0.9942$^{*}$ &    \hspace{0.25em}0.9377 &    \hspace{0.25em}0.2663 &  N/A \\
VREx      &        \hspace{0.25em}0.9688 &    \hspace{0.25em}0.9866 &   \hspace{0.25em}0.9787 &    \hspace{0.25em}0.9851 &    \hspace{0.25em}0.5169 &  N/A \\
\bottomrule
\end{tabular}
    }
    \label{tab:full_correlation_results_coloredmnist}
\end{table}

\begin{table}[h]
    \centering
    \caption{Correlations between probing results and generalization accuracies, on VLCS}
    \resizebox{0.8\linewidth}{!}{
    \begin{tabular}{lllllll}
\toprule
Algorithm & Probe\_0 & Probe\_1 & Probe\_2 & Probe\_3 & Probe\_4 & Probe\_5 \\
\midrule
ANDMask   &                  -0.7884 &                -0.6042 &                 -0.8627 &                -0.7988 &                -0.6038 &                -0.9108$^{*}$ \\
CAD       &                     -0.8111 &                -0.6155 &                 -0.6242 &                -0.7604 &                -0.7486 &                -0.9077$^{*}$ \\
CDANN     &       \hspace{0.25em}0.0623 &                -0.5973 &                 -0.3463 &  \hspace{0.25em}0.5724 &   \hspace{0.25em}0.021 &   \hspace{0.25em}0.5191 \\
CORAL     &                     -0.6874 &                -0.3425 &                 -0.4622 &                -0.6697 &                -0.7194 &                 -0.6265 \\
CondCAD   &                    -0.9467$^{*}$ &                -0.8162 &                  -0.863 &               -0.967$^{**}$ &               -0.9185$^{*}$ &              -0.9941$^{***}$ \\
DANN      &                     -0.1123 &                 -0.629 &                 -0.5488 &                -0.6396 &                -0.3297 &   \hspace{0.25em}0.4269 \\
ERM       &                     -0.8021 &                -0.4361 &                 -0.7221 &                -0.8069 &                -0.8837 &                 -0.8699 \\
GroupDRO  &                     -0.8521 &                -0.7036 &                 -0.7736 &                -0.8923 &               -0.9307$^{*}$ &                -0.9052$^{*}$ \\
IB\_ERM    &                     -0.8316 &                -0.5454 &                 -0.7996 &                -0.8347 &                -0.8475 &                 -0.8376 \\
IB\_IRM    &                   -0.9672$^{**}$ &               -0.9477$^{*}$ &                -0.9136$^{*}$ &             -0.9915$^{***}$ &             -0.9974$^{***}$ &                 -0.7454 \\
IRM       &                   -0.9979$^{**}$ &                -0.8173 &                  -0.943 &                -0.9651 &                 -0.956 &    \hspace{0.25em}0.445 \\
MLDG      &                     -0.7728 &                -0.6106 &                 -0.6124 &                -0.8167 &                -0.8326 &                  -0.821 \\
MMD       &                     -0.7461 &                -0.6395 &                 -0.5214 &                -0.7459 &                -0.8597 &                  -0.716 \\
MTL       &                     -0.8281 &                -0.8388 &                 -0.7541 &                 -0.711 &                -0.6164 &                 -0.7218 \\
Mixup     &                     -0.6426 &                -0.5704 &                 -0.5358 &                -0.6617 &                -0.7821 &                 -0.7691 \\
RSC       &                     -0.7308 &                -0.4961 &                 -0.5818 &                -0.8237 &               -0.9211$^{*}$ &                -0.9132$^{*}$ \\
%SANDMask  &                     -0.6535 &                -0.6089 &                 -0.6556 &                -0.7391 &                -0.8462 &                 -0.8382 \\
SD        &                     -0.6835 &                -0.3245 &                 -0.6887 &                -0.7215 &                -0.7497 &                 -0.7583 \\
SagNet    &                     -0.6156 &                -0.8272 &                 -0.7361 &                -0.7811 &                -0.8167 &                -0.9139$^{*}$ \\
SelfReg   &                       -0.85 &                -0.7936 &                 -0.7676 &                -0.7544 &                -0.6486 &                 -0.7837 \\
TRM       &       \hspace{0.25em}0.5303 &  \hspace{0.25em}0.8719 &  \hspace{0.25em}0.9031$^{*}$ &  \hspace{0.25em}0.8972 &  \hspace{0.25em}0.8852 &  \hspace{0.25em}0.9105$^{*}$ \\
VREx      &                     -0.7578 &                -0.6201 &                 -0.7837 &                -0.7981 &                 -0.762 &                 -0.8093 \\
\bottomrule
\end{tabular}
    }
    \label{tab:full_correlation_results_VLCS}
\end{table}

\begin{table}[h]
    \centering
    \caption{Correlations between probing results and generalization accuracies, on PACS}
    \resizebox{0.8\linewidth}{!}{
    \begin{tabular}{lllllll}
\toprule
Algorithm & Probe\_0 & Probe\_1 & Probe\_2 & Probe\_3 & Probe\_4 & Probe\_5 \\
\midrule
ANDMask   &    \hspace{0.25em}0.7009 &   \hspace{0.25em}0.6549 &    \hspace{0.25em}0.7471 &    \hspace{0.25em}0.7242 &    \hspace{0.25em}0.7655 &  \hspace{0.25em}0.9147$^{*}$ \\
CAD       &       \hspace{0.25em}0.7815 &  \hspace{0.25em}0.9024$^{*}$ &    \hspace{0.25em}0.7055 &     \hspace{0.25em}0.854 &    \hspace{0.25em}0.6974 &   \hspace{0.25em}0.3663 \\
CDANN     &                    -0.9258$^{*}$ &                -0.9026$^{*}$ &                  -0.7547 &    \hspace{0.25em}0.8968 &    \hspace{0.25em}0.4159 &   \hspace{0.25em}0.7199 \\
CORAL     &       \hspace{0.25em}0.6937 &   \hspace{0.25em}0.6337 &    \hspace{0.25em}0.8343 &    \hspace{0.25em}0.7439 &   \hspace{0.25em}0.9382$^{*}$ &   \hspace{0.25em}0.7638 \\
CondCAD   &        \hspace{0.25em}0.661 &   \hspace{0.25em}0.7221 &    \hspace{0.25em}0.7567 &    \hspace{0.25em}0.7183 &    \hspace{0.25em}0.7373 &  \hspace{0.25em}0.9398$^{*}$ \\
DANN      &       \hspace{0.25em}0.1647 &   \hspace{0.25em}0.2556 &    \hspace{0.25em}0.0702 &                   -0.202 &    \hspace{0.25em}0.2035 &   \hspace{0.25em}0.0368 \\
ERM       &       \hspace{0.25em}0.4843 &   \hspace{0.25em}0.5695 &    \hspace{0.25em}0.6677 &    \hspace{0.25em}0.6186 &    \hspace{0.25em}0.4791 &   \hspace{0.25em}0.6495 \\
GroupDRO  &       \hspace{0.25em}0.5355 &   \hspace{0.25em}0.6534 &    \hspace{0.25em}0.6629 &     \hspace{0.25em}0.641 &    \hspace{0.25em}0.6643 &    \hspace{0.25em}0.665 \\
IB\_ERM    &        \hspace{0.25em}0.741 &   \hspace{0.25em}0.8657 &    \hspace{0.25em}0.8592 &    \hspace{0.25em}0.7541 &    \hspace{0.25em}0.7057 &   \hspace{0.25em}0.6374 \\
IB\_IRM    &       \hspace{0.25em}0.7409 &   \hspace{0.25em}0.6835 &      \hspace{0.25em}0.71 &    \hspace{0.25em}0.8454 &    \hspace{0.25em}0.8313 &   \hspace{0.25em}0.8211 \\
IRM       &       \hspace{0.25em}0.8565 &  \hspace{0.25em}0.9172$^{*}$ &  \hspace{0.25em}0.9833$^{**}$ &  \hspace{0.25em}0.9654$^{**}$ &  \hspace{0.25em}0.9758$^{**}$ &  \hspace{0.25em}0.9398$^{*}$ \\
MLDG      &       \hspace{0.25em}0.8648 &  \hspace{0.25em}0.9347$^{*}$ &  \hspace{0.25em}0.9774$^{**}$ &    \hspace{0.25em}0.8492 &   \hspace{0.25em}0.9193$^{*}$ &  \hspace{0.25em}0.9298$^{*}$ \\
MMD       &       \hspace{0.25em}0.5435 &   \hspace{0.25em}0.6747 &    \hspace{0.25em}0.6089 &     \hspace{0.25em}0.529 &    \hspace{0.25em}0.6636 &   \hspace{0.25em}0.6518 \\
MTL       &       \hspace{0.25em}0.5355 &   \hspace{0.25em}0.7426 &    \hspace{0.25em}0.6701 &    \hspace{0.25em}0.6356 &    \hspace{0.25em}0.6907 &   \hspace{0.25em}0.7749 \\
Mixup     &       \hspace{0.25em}0.7521 &   \hspace{0.25em}0.7421 &    \hspace{0.25em}0.8298 &    \hspace{0.25em}0.6567 &    \hspace{0.25em}0.8161 &   \hspace{0.25em}0.8569 \\
RSC       &       \hspace{0.25em}0.5571 &   \hspace{0.25em}0.6241 &    \hspace{0.25em}0.7671 &    \hspace{0.25em}0.6141 &    \hspace{0.25em}0.6833 &   \hspace{0.25em}0.0018 \\
%SANDMask  &        \hspace{0.25em}0.135 &   \hspace{0.25em}0.0436 &      \hspace{0.25em}0.16 &    \hspace{0.25em}0.2494 &    \hspace{0.25em}0.3938 &   \hspace{0.25em}0.3387 \\
SD        &       \hspace{0.25em}0.4503 &   \hspace{0.25em}0.4382 &    \hspace{0.25em}0.7056 &    \hspace{0.25em}0.5861 &    \hspace{0.25em}0.5235 &   \hspace{0.25em}0.5208 \\
SagNet    &       \hspace{0.25em}0.2259 &   \hspace{0.25em}0.2972 &    \hspace{0.25em}0.3456 &    \hspace{0.25em}0.1986 &                  -0.8632 &                 -0.7221 \\
SelfReg   &       \hspace{0.25em}0.4667 &   \hspace{0.25em}0.6937 &     \hspace{0.25em}0.772 &    \hspace{0.25em}0.7706 &    \hspace{0.25em}0.6388 &   \hspace{0.25em}0.8693 \\
TRM       &       \hspace{0.25em}0.3149 &   \hspace{0.25em}0.6426 &    \hspace{0.25em}0.5857 &    \hspace{0.25em}0.6102 &    \hspace{0.25em}0.5043 &   \hspace{0.25em}0.3968 \\
VREx      &                     -0.1246 &   \hspace{0.25em}0.0428 &    \hspace{0.25em}0.0869 &                  -0.0261 &                    -0.08 &   \hspace{0.25em}0.1654 \\
\bottomrule
\end{tabular}
    }
    \label{tab:full_correlation_results_PACS}
\end{table}

%%%%%%%%%%%%%%%%%%%%%%%%%%%%%%%%%%%%%%%%%%%%%%%%%%%%%%%%%%%%%%%%%%%%%%%%%%%%%%%
%%%%%%%%%%%%%%%%%%%%%%%%%%%%%%%%%%%%%%%%%%%%%%%%%%%%%%%%%%%%%%%%%%%%%%%%%%%%%%%

\end{document}